\documentclass[letterpaper]{article} % DO NOT CHANGE THIS
\usepackage{aaai25}  % DO NOT CHANGE THIS
\usepackage{graphicx} % DO NOT CHANGE THIS
\usepackage{natbib}  % DO NOT CHANGE THIS AND DO NOT ADD ANY OPTIONS TO IT
\usepackage{caption} % DO NOT CHANGE THIS AND DO NOT ADD ANY OPTIONS TO IT
\usepackage{algorithm}
\usepackage{algorithmic}

\usepackage{tikz}
\usetikzlibrary{arrows.meta} % for advanced arrow tips in TikZ

\usepackage{fontawesome5} % For icons like user, robot, chess-knight

\usepackage{newfloat}
\usepackage{listings}
\DeclareCaptionStyle{ruled}{labelfont=normalfont,labelsep=colon,strut=off} % DO NOT CHANGE THIS
\floatstyle{ruled}
\newfloat{listing}{tb}{lst}{}
\floatname{listing}{Listing}
\newtheorem{definition}{Definition}

\usetikzlibrary{positioning, shapes.geometric, arrows.meta, shadows}

\begin{document}
\title{Outer-Learning Framework for Playing Multi-Player Trick-Taking Card Games: A Case Study in Skat
}

%%% Provide names, affiliations, and email addresses for all authors.

\author{Stefan Edelkamp \\ 
Charles University \\
Prague, Czech Republic \\
\texttt{stefan.edelkamp@mff.ktiml.cuni.cz}}

%\author{Rainer G\"o{\ss}l}
%\affiliation{
%  \institution{}
%  \city{Chemnitz}
%  \country{Germany}}
%email{rgoessl@aol.de}

%%% Use this environment to specify a short abstract for your paper.
%\author{Stefan Edelkamp}
%\date{June 2025}

\maketitle

\begin{abstract}
In multi-player card games such as Skat or Bridge, the early stages of the game, such as bidding, game selection, and initial card selection, are often more critical to the success of the play than refined middle- and end-game play. At the current limits of computation, such early decision-making resorts to using statistical information derived from a large corpus of human expert games. In this paper, we derive and evaluate a general bootstrapping outer-learning framework that improves prediction accuracy by expanding the database of human games with millions of self-playing AI games to generate and merge statistics. We implement perfect feature hash functions to address compacted tables, producing a self-improving card game engine, where newly inferred knowledge is continuously improved during self-learning. The case study in Skat shows that the automated approach can be used to support various decisions in the game. 
%Multi-agent card games are testbeds for modern AIs playing at the edge of human strength. In popular trick-taking games, such as Skat or Bridge, the early stages of the game, such as bidding, game selection, and initial card selection, are often more critical to the success of the play than refined middle- and end-game play. At the current limits of computation, such early decision-making resorts to using statistical information derived from a large corpus of human expert games, e.g., to predict winning probabilities. 
%In this paper, we derive and evaluate a general bootstrapping outer-learning framework that improves prediction accuracy by expanding the database of human games with millions of self-playing AI games to generate, merge, and improve statistics. We show that the automated approach can be used to support various decisions in the game. We conduct a case study in Skat, competitively played in national leagues and international championships. We implement perfect feature hash functions to address compacted tables, producing a self-improving card game engine, where newly inferred predictions were improved continuously during self-learning. 
\end{abstract}

\section{Introduction}

In recent years, significant progress has been made in automated playing fully observable tabletop games such as Go~\cite{AlphaGo} or Chess~\cite{alphazero}, 
and partially observable two-player games such as Poker~\cite{poker} or Stratego~\cite{doi:10.1126/science.add4679}, playing
beyond the level of human strength.

Despite some success stories~\cite{stone2024setbasedretrogradeanalysisprecomputing,DBLP:conf/ijcai/LiZCV22,DBLP:conf/cg/Edelkamp22}, the supremacy race between humans and computers to play complex multi-player trick-taking card games such as Bridge~\cite{GIB,DBLP:journals/corr/abs-1911-07960} or Skat~\cite{kupferschmid:masters-thesis-03,DBLP:conf/ijcai/BuroLFS09,DBLP:journals/corr/abs-1903-09604} is still open. 
This observation is especially true for playing the initial stages of the games, namely bidding, game selection, or card openings. 

These card games are very difficult to tackle because of the large number of initial hands together with the wide horizon of play. This makes it difficult to backpropagate or store the outcome of the game at the start of the game, for example, by set-based retrograde analysis, ~\cite{stone2024setbasedretrogradeanalysisprecomputing}, or by minimizing counterfactual regret~\cite{DBLP:journals/corr/abs-1811-00164}. 

For imperfect information games look-ahead search on top of policy networks~\cite{Lookahead},  value functions for depth-limited solving in zero-sum imperfect information games~\cite{depthlimited},
opponent-limited online search~\cite{Opponent}, and a unified learning algorithm~\cite{studentgames} have been proposed.
While there are advances in metalearning for general sum games \cite{sychrovský2025approximatingnashequilibriageneralsum}, for the initial stages of general card games, 
research resorts to approximating the winning probabilities to decide on which game to play
and which opening card to choose~\cite{DBLP:conf/socs/Edelkamp19}. 

Building and training a
neural network to predict the bidding odds and the next card to play in a complex game like Bridge and Skat is a demanding task, which for highly advanced play is apparently challenging current technology~\cite{DBLP:journals/corr/abs-1903-00900,rebstock2024transformerbasedplanningobservation,rebstock2019learningpolicieshumandata}. 
For the initial stages of games, such as bidding, selecting the game to declare, or discarding cards, due to the length of the trick-tacking stage, standard machine learning 
often does not provide sufficient guidance~\cite{biddingskat,DBLP:conf/ecai/CohensiusMOS20}. While open card games are simply to classify using classical game-tree search algorithms, they do not provide substantial insights to the game, as there are many moves based on the incomplete information the players have. Monte-Carlo sampling of the incomplete information does not necessarily compensate for the lack of information moves~\cite{6203567}.

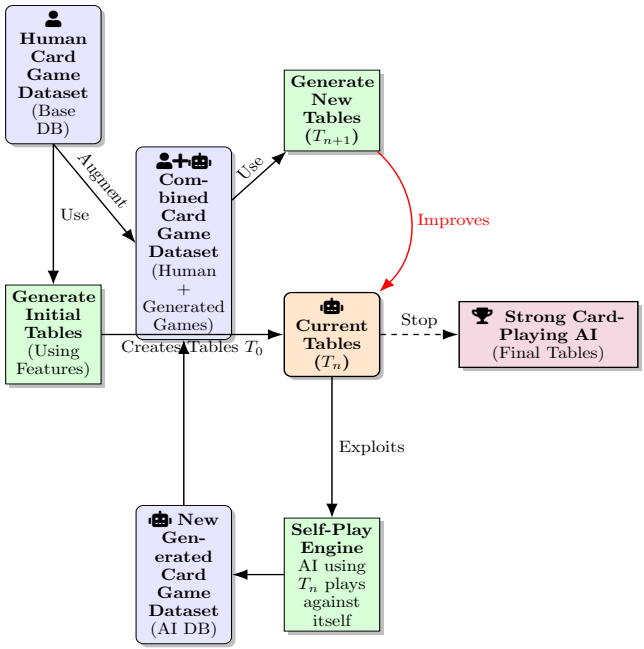
\begin{figure}[t!]
    \centering 
\resizebox{8.5cm}{8.5cm}{%
\begin{tikzpicture}[
    scale=0.25,
    node distance=3cm and 1cm,
    % Define styles for different types of nodes
    data/.style={
        rectangle, 
        draw, 
        rounded corners, 
        fill=blue!10, 
        minimum width=1.75cm, 
        minimum height=0.75cm,
        text centered,
        text width=1.6cm,
        drop shadow
    },
    process/.style={
        rectangle, 
        draw, 
        fill=green!15, 
        minimum width=1.75cm, 
        minimum height=0.75cm,
        text centered,
        text width=1.6cm,
        drop shadow
    },
    model/.style={
        rectangle, 
        draw, 
        thick,
        rounded corners, 
        fill=orange!20, 
        minimum width=1.75cm, 
        minimum height=0.75cm,
        text centered,
        text width=1.6cm,
        drop shadow
    },
    io/.style={ % For the final output
        rectangle,
        draw,
        thick,
        fill=purple!15,
        minimum width=1.75cm, 
        minimum height=0.75cm,
        text centered,
        text width=3.2cm,
        drop shadow
    },
    % Style for arrows
    arrow/.style={
        -{Latex[length=3mm]},
        thick
    },
    % Style for the loop annotation box
    loop/.style={
        draw,
        dashed,
        rounded corners,
        line width=0.8pt,
        inner sep=0.7cm,
        label={[font=\bfseries]above:Bootstrap Iteration Loop}
    }
]

% --- NODES ---

% 1. Initial Human Data
\node (humandata) [data] {
    \faUser{} 
    \textbf{Human Card Game Dataset} \\
    (Base DB)
};

% 2. Initial Training Process
\node (train0) [process, below=of humandata] {
    \textbf{Generate Initial Tables} \\
    (Using Features)
};

% 3. The Current Model inside the loop (Version n)
\node (modeln) [model, right=of train0, xshift=2.5cm] {
    \faRobot{} 
    \textbf{Current Tables ($T_n$)} 
    %\\
%    (e.g., Policy + Value Network)
};

% 4. Self-Play Process
\node (selfplay) [process, below=of modeln] {
    \textbf{Self-Play Engine} \\
    AI using $T_n$ 
    plays against itself
};

% 5. New Computer-Generated Data
\node (computerdata) [data, left=of selfplay] {
    \faRobot{} 
    \textbf{New Generated Card Game Dataset} \\
    (AI DB)
};

% 7. Training Process for the next model
\node (trainn) [process, above=of modeln] {
    \textbf{Generate New Tables ($T_{n+1}$)} \\
%    (Reinforcement Learning)
};

% 6. Combined Dataset
\node (combineddata) [data, below left = -0.1cm and 1cm of trainn] {
    \faUser\faPlus\faRobot{} 
    \textbf{Combined Card Game Dataset} \\
    (Human + Generated Games)
};

% 8. Final Output
\node (finalai) [io, right=of modeln, xshift=0.5cm] {
    \faTrophy{} \
    \textbf{Strong Card-Playing AI} \\
    (Final Tables)
};

\draw [arrow] (humandata) -- (train0) node[midway, right] {Use};
\draw [arrow] (train0) -- (modeln) 
node[midway, below] {Creates Tables $T_0$};

% The main bootstrap loop
\draw [arrow] (modeln) -- (selfplay) node[midway, right] {Exploits};
\draw [arrow] (selfplay) -- (computerdata);

% --- ARROWS ---

% Initial flow from human data to the first model version

% Combining data
\draw [arrow] (computerdata) -- (combineddata);
\draw [arrow] (humandata.south) -- % ++(0,-2.5) -| 
(combineddata.west) node[pos=0.45, above, sloped] {Augment};

% Training the new model and feedback loop
\draw [arrow] (combineddata) -- (trainn) node[midway, above, sloped] {Use};
\draw [arrow, red] (trainn) to [bend left=45] node[midway, right] {Improves} (modeln);

% Final output
\draw [arrow, dashed] (modeln) -- (finalai) node[midway, above] {Stop};
%\node at ($(modeln)!0.5!(finalai)$) [above] {iterations};

% --- ANNOTATION ---

% Draw a dashed box around the iterative loop
%\node[loop, fit=(modeln) (selfplay) (computerdata) (combineddata) (trainn)] (loop_box) {};

\end{tikzpicture}
}
    \caption{Outer-learning framework for trick-taking card games. Tables are generated by processing Human and AI games, where the set Human games form the initial base table. The AI generates new games during self-play to construct new tables and merging them to the existing ones. Tables entries compress information based on selected features and contain winning probabilities or card recommendations. }
    \label{fig:outerframework}
\end{figure}

In this paper, we propose a novel and generic self-learning bootstrapping technique without employing neural networks (see the architecture illustration in Fig.~\ref{fig:outerframework}), which we call \emph{outer-learning procedure} for
large game datasets (in contrast to \emph{inner-learning procedure} for the representation and progression of knowledge within the players). The approach enlarges the statistical base tables used as decision support in the algorithms with the outcome of playing a large body of games. 

The paper is structured as follows. First, we look at related work and introduce briefly the game of Skat.
Next, we propose a portable game notation for card games that we adopted and extended from the chess community, and explain how to construct the so-called winning tables. We study three different cases for implementing the general self-learning procedure in the game of Skat. Experiments show
how the automated procedure improves the AI game engines' playing quality, measured in the accuracy of matching the
predicted result of the open-card solver.
We end with some concluding remarks.

\section{Related Work}

Perfect-information Monte-Carlo sampling (PIMC) still is a state-of-the-art algorithm for imperfect information games and was used in Ginsberg's GIB \cite{GIB} and other trick-taking card games. The reasons why PIMC was successful have been analyzed by~\cite{skatlong}.
However, PIMC has issues with
strategy fusion and non-locality leading to the $\alpha\mu$ algorithm \cite{DBLP:journals/corr/abs-1911-07960}. For mere trick-taking, promising results even against Bridge champions were obtained\footnote{https://www.imperial.ac.uk/news/235238/ai-based-imperial-research-beats-world}, but for the bidding and game selection stages, the algorithm was not used.

Self-learning has been very effective for deep reinforcement learning in board games.  AlphaGo~\cite{AlphaGo} was trained on a large corpus of human games to learn how to play. AlphaZero~\cite{alphazero} improved upon this by playing against itself; in the extreme case starting from completely random play.
While AlphaZero exploited deep neural networks together with Monte Carlo tree search for exploration, we use winning features and statistics to derive accurate winning probabilities and card recommender systems for game openings.

There are visible successes in deep learning for playing two-player zero-sum incomplete information board games like Stratego~\cite{Perolat_2022}. 
The regularized Nash dynamics (RNaD) algorithm uses policy-gradient methods in imperfect
information games and has been applied to general games like the two-layer simplified card game Goofspiel~\cite{DBLP:conf/ijcai/KubicekBL24}, but for multi-player card games that include bidding, trick-taking, and scoring, the task is more
challenging. 

There are (moderately strong)
neural network players for Talon-based games, such as Mahjong~\cite{DBLP:journals/corr/abs-1906-02146}, but the best AI players in competitive Bridge and Skat
are mainly combinatorial algorithms~\cite{skatfurtak,DBLP:conf/ijcai/BuroLFS09}:
partial-ordering
minimax and $\alpha\mu$~\cite{DBLP:conf/ijcai/LiZCV22,isaim,DBLP:conf/ijcai/LiZCV22},
extensive retrograde open card analyses for PIMC~\cite{stone2024setbasedretrogradeanalysisprecomputing}, as well as refined search procedures~\cite{DBLP:journals/corr/abs-1905-10911, DBLP:journals/corr/abs-1905-10907,DBLP:conf/cig/Edelkamp21}. Large-language models 
remain to be improved 
to play competitively in such complex card games ~\cite{wang2025largelanguagemodelsmaster}.

Similarly to this line of research, some works~\cite{DBLP:conf/socs/Edelkamp19,DBLP:journals/corr/abs-1905-10907} recorded a larger number of amateur and professional human card games to elicit knowledge from them. A refined selection of winning parameters was exploited to exploit statistical information on the outcome of the game in tables.
Unified self-play and solution pipelines with theoretical guarantees were given by \cite{studentgames},
and large-scale self-play successes were reported in the partially observable two-player zero-sum game of Stratego~\cite{Perolat_2022}. In contrast with these findings, we highlight our advances in implicitity, generality, and latency, while we acknowledge limitations in theoretical guarantees.
 
\subsection{Skat}

Skat\footnote{The exact rules of Skat are involved to be replicated in this
work, so that we refer the interested reader to https://www.pagat.com/schafkopf/skat.html} is the German national card game with international reach and an estimated number of several million players. 
Competitive Skat is defined by the International Skat Players Association (ISPA)\footnote{see http://www.ispaworld.info}.
The highest level Skat competition in the world is the ISPA World Skat Championship, played biannually since 1978 at various locations worldwide.  It alternates with the ISPA European Skat Championship, played biannually since 1979. 
There are several leagues played in Skat, including the 1st and the 2nd German Bundesliga. There even is an International Skat Court in Augsburg as highest decision-making body that oversees the observance of the regulations for Skat, the refereeing regulations, and the rules for referees. 
In short, Skat has four different playing stages.

\begin{description}
\item [bidding stage] Players are trying to become a declarer, announce and commit to bits.
\item [skat-putting stage (optional)] The declarer takes and discards two skat cards.
\item [game-selection stage] The declarer announces the type of game to be played and (rarely) an increased contract. 
\item [trick-taking stage] By placing the cards on the table in turns, the winner of the trick continues with the next trick.
\end{description}
 
At the beginning of a game, each player gets 10 cards, which are hidden to the other players. The remaining two cards, called skat cards, are placed face down on the table. 
%Each hand is played in two stages, bidding and card play. 

\begin{figure}[t]
    \centering
    \includegraphics[width=9cm]{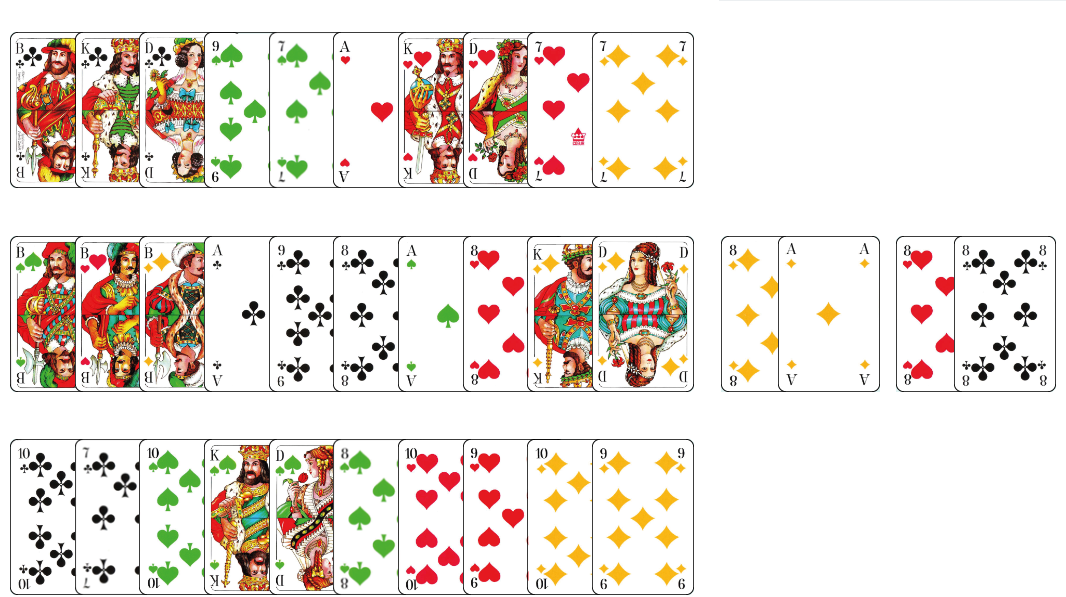}
    \caption{Skat deal with Diamond 8 and Ace being taken, and Heart \& Club 8 being discarded by the declarer in the middle. }
    \label{fig:skatput}
\end{figure}

The bidding stage determines the declarer and the two opponents. Two players announce and accept increasing bids until one passes. 
The winner of the first bidding phase continues bidding with the third player. The successful bidder in the second bidding phase plays against the other two. The maximum bid a player can announce depends on the type of game the player wants to play, and, in case of a trump game, a multiplication factor determined by the jacks. The declarer decides on the game to be played. Before declaring, he may pick up the skat cards and then discard two cards from his hand, face down. These cards count towards the declarer's score. 
%The declarer selects and discards two of 12 cards from his/her hand to be put. 
Similarly to choosing the game for which to bid and to declare afterwards, selecting good skat cards is crucial, as an unfortunate choice often leads to a loss in a game that could have been won with a better one. All these first stages are based on accurately estimating the winning probabilities derived from a corpus of expert games.

\begin{figure}
    \centering
    \includegraphics[width=1\linewidth]{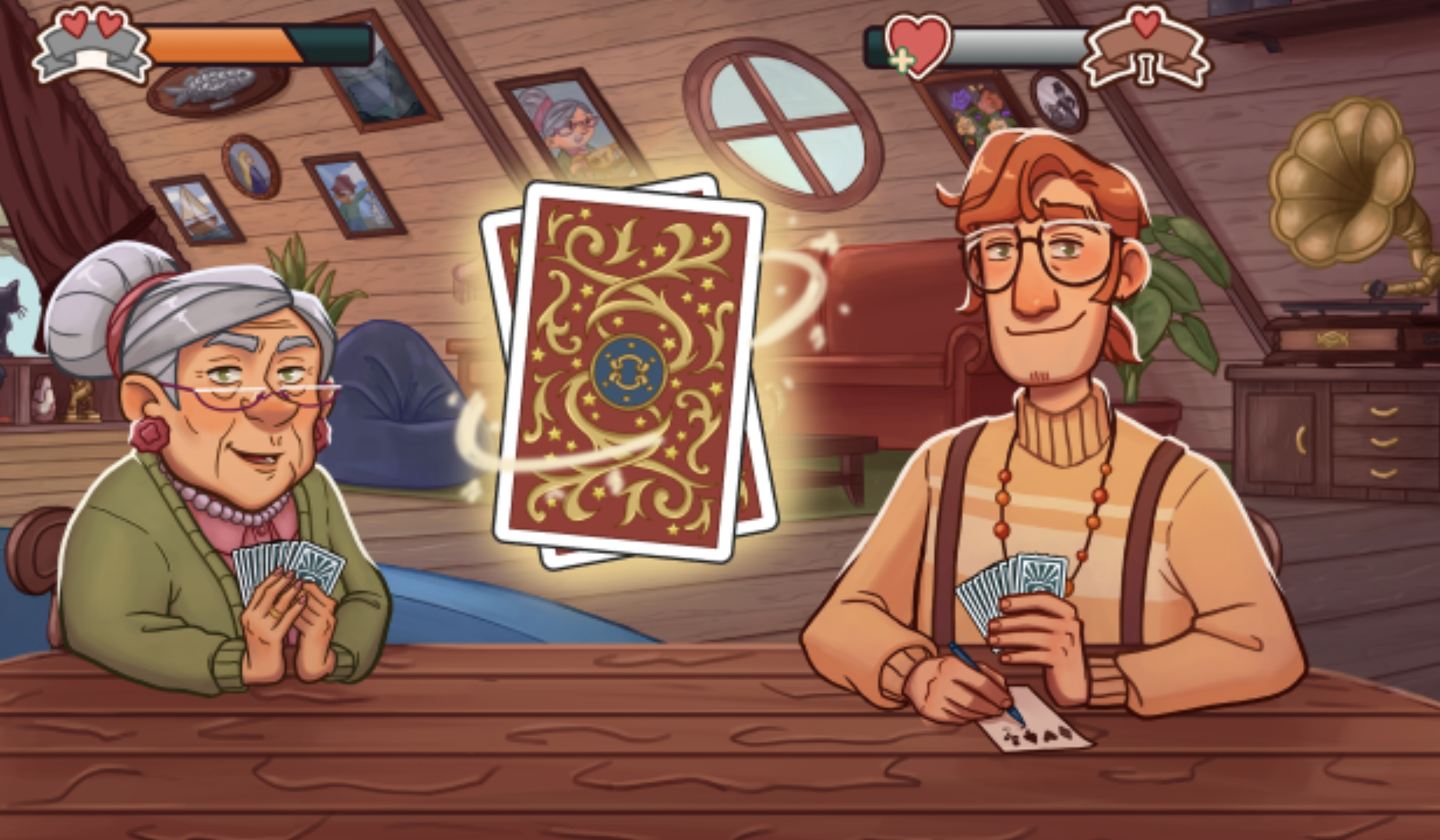}
\includegraphics[width=1\linewidth]{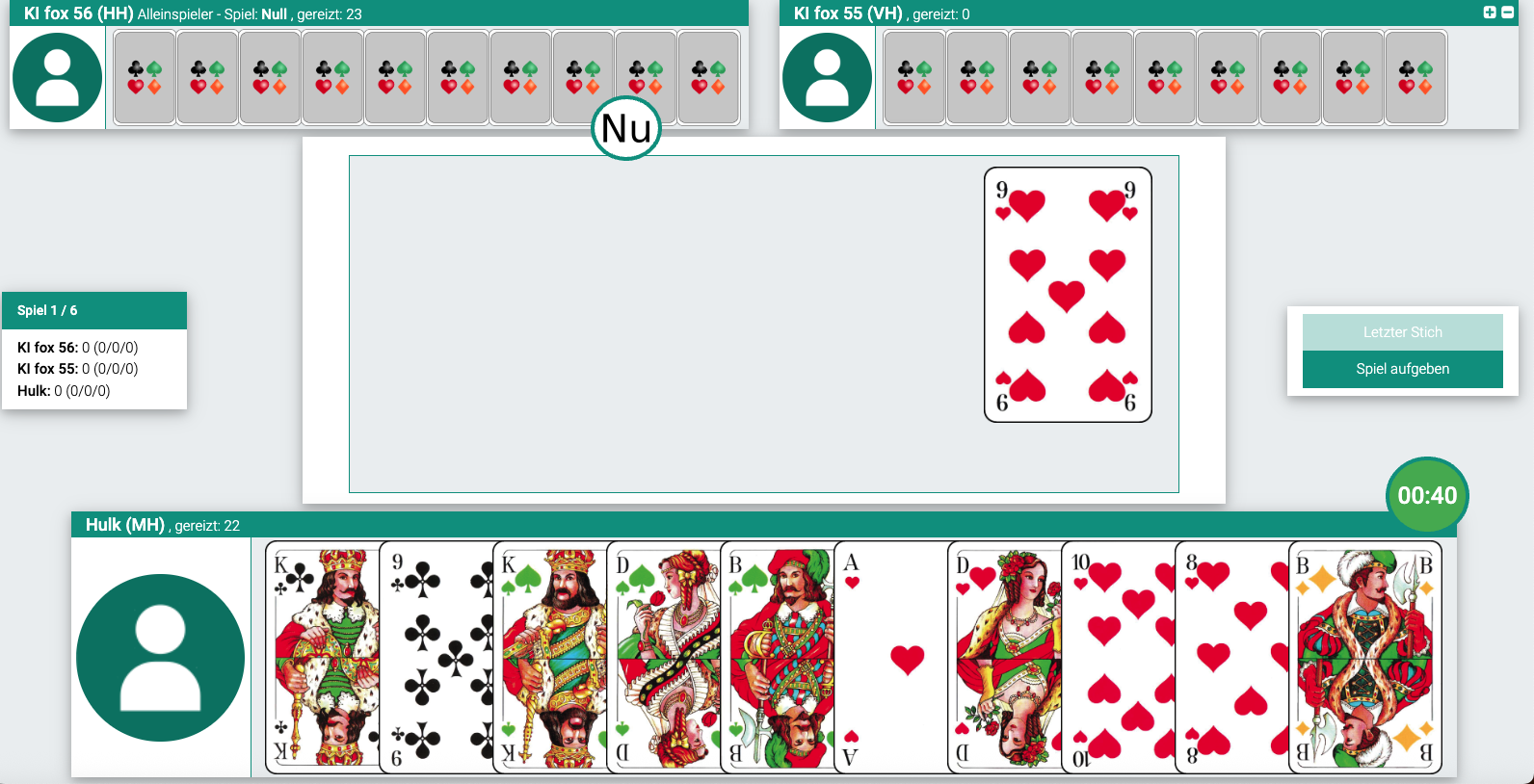}
    \caption{A single-mobile Skat player application (top, see https://youtu.be/UkubGNNF2PY), and a multi-player Skat web application 
    (bottom, see www.skat-ki.de).}
  \label{fig:bestjack}
\end{figure}

Fig.~\ref{fig:skatput} provides a small example of the Skat game at the start of the trick-taking stage. 
All 32 cards of the deal are shown, while the players only see their individual cards, with the declarer also knowing the skat cards (if taken).
Skat AIs are desired to serve as a basis for mobile and server player  applications (e.g., see Fig.~\ref{fig:bestjack}).

\section{Constructing Winning Tables}

Edelkamp~\cite{10.1007/978-3-031-70893-0_6} has generalized the implementation of a Skat player to a number of different two-team trick-taking games, including Belote, Schafkopf, Euchre, Tarot (four-player game variant), Doppelkopf (two variants, including ones with double decks of $2 \times 20$ cards and $2 \times 24$ cards), and Spades (often played in the US).  

Let $p$ be the number of players,
$n$ be the size of the deck, then $\lfloor n/p \rfloor$ is the initial number of cards per hand. Neglecting the initial turn, the number of possible deals is equal to
$\prod_{i=1}^p {{n - i \cdot \lfloor n/p \rfloor} \choose \lfloor n/p \rfloor}$.
 
\begin{definition} [Trick-Taking Game] 
A \emph{trick-taking card game} ${\mathcal G} = \langle {\mathcal C},p,c,r\rangle$ of $p$ players
with a deck of $n=|{\mathcal C}|$ cards in a (multi) set ${\mathcal C}$ is played in several rounds, 
called tricks $t\in T \subseteq {\mathcal C}^p$, which are each evaluated to determine a \emph{winner} %$: T \rightarrow \{0,1\}$ 
of that trick.
In the view of each player, there are $l$ teams of players, so that %the team
membership can be realized as a mapping $\mbox{\em team} :\{1,\ldots,p\} \rightarrow \{0,1,\ldots,l\}$.
Each trick induces a cost $c(t)$ defined by the rules $r$ of the games.
Trick costs are assigned to the trick winning team. 
\end{definition}

Although we stick to Skat and $l=2$, the concept of winning tables can be used for all the above games and beyond. The main reason we did not extend the approach to other games  is that we have not extracted compelling  features, nor access to large body of base tables. 

For the approach to work, we start with a decision making alias choice question $Q$, which we aim to answer by exploiting the winning statistics derived from a large set of played games $G$, to support the AI card game playing engine in the early stages of the game. 

As for early play, applying game theory with Nash equilibria or backward analyzes like 
regret minimization currently seem either  inaccurate or computationally infeasible, and open-card solving does not
cover well the amount of incomplete information and the amount of knowledge exchange to reduce it. 
The main reason is that trick-taking card games feature a large amount of private information that slowly becomes revealed through a long sequence of actions. This makes the number of histories exponentially large in the action sequence length, as well as creating extremely large information sets. Together with the huge number of possible deals, the incomplete-information card games mentioned above are too large.
If at each point in the game. we were to efficiently compute the exact winning probability of each valid next card choice in the hand the players (w.r.t. the entire history of the game), the card game problem could be considered solved. We would always play the card that offers the highest probability of winning, and with a binary search we could find the card with the maximal score, as the recommended card should result in the best possible expected score, weighting in all uncertainty. 
However, due to the large number of deals and information sets together with all possible bidding and playing sequences, solving is beyond current research.

Hence, we stick to approximating probabilities. In the Skat game with $3 \times 32!/10!10!10!2!$ deals (factor 3 accounting for three different forehand positions), questions $Q$ could correspond to \emph{what is the winning probability of a given game type and bidding level for the 12 cards before/after Skat taking/putting?} or
\emph{What card should be chosen to open the game as opponent?} To derive a continuous self-learning and bootstrapping framework for answering such questions on growing corpora of games, we start with a few definitions.

\begin{definition} (Game Database)
    Let $g$ be one card game that includes information about all stages of the games (e.g., deal, bidding, game declaration, trick taking, outcome). Let \emph{game database} $G$ include all games $g$ that have been played.
\end{definition}
 
We store the games in our proposal of a \emph{portable game notation} (PGN). Unlike games like chess, we record series of games, with each line corresponding to one game. The series can be merged into a database of games.
In the header, we find the player names, the data and time of the series, and the statistics in terms of the score and the winning ratios. Fig.~\ref{fig:truncatedpgn} shows one (fragmented) example.

\begin{figure}[t]
\begin{tiny}
\begin{verbatim}
[Event: Training] [Date: 12.01.2025, 13:19, Duration: 24m] 
[Seat1: Human (ID: 23)] [Seat2: AI1 (ID: 68)] [Seat3: AI2 (ID: 67)] 
[Number of games: 36] 
[Result: Seat1: 581 (6:3) Seat2: 1276 (9:0) Seat3: 1136 (13:3)] 

0 0 2 22 0 22 0 0 0 0 0 0 1 0 -1 87 10 ... 
...
\end{verbatim}
\end{tiny}
\caption{Truncated example for a game series in PGN.}
\label{fig:truncatedpgn}
\end{figure}

The information stored for each game is game dependend and follows a certain scheme (see Table~\ref{tab:pgn} for the Game of Skat). As trick-taking can end early (e.g., by the declarer showing cards or opponents folding), the length of a game may vary. With a game in PGN, a game can be reconstructed, annotated, and debugged. As it is of limited impact even for advanced play, the PGN does not yet reflect the (for Skat quite complex) dynamics of the bidding stage, but only the final values called at each player position. Timing information is also neglected in the PGN, whereas temporal signals such as hesitation are actively used in top play to indicate deviations from standard playing conventions. Both data sources are available and can be included for an extended PGN for professional play.

To answer a question $Q$, we need to find statistically relevant winning features and, at the same time, to ensure that the cross product of their domains remains moderately small, so that discrete tables can be built and addressed on a given computer device. 
     
\begin{definition} (Relevant Winning Feature)
    Let $g$ be one card game in $G$ that includes the information about all stages of the games (e.g. deal, bidding, game declaration, trick taking), and for $i \in \{1,\ldots,k\}$ let $f_i(g)$ be a \emph{relevant feature/parameter} of finite domain $D_i$, and $F(g) = (f_1(g),\ldots,f_k(g))
    \in D_1 \times \ldots \times D_k$ be the vector of all relevant winning features.
\end{definition}

Determining the relevant winning features for a given game and question $Q$ involves analyzing a wide range of statistical game data. (For Skat this work has been done by G\"o{\ss}l and documented in Chapter 3 of his book~\cite{Rainer}\footnote{see
https://www.skatfuchs.eu/SB-Kapitel3.pdf
}, so that we took the winning features from there.) Hash functions
map feature vectors of games 
to a table.

\begin{definition}[Bucket]
A \emph{bucket} is a set of games that share the same hash value of their selected features. 
\end{definition}

We hash the games into buckets of common feature values to derive winning statistics in each of the buckets.
The games themselves are not stored.
For the sake of efficiency in time and space, we use perfect hash functions~\cite{phf,dphf}. 
Recall that perfect hashing and their inverses have been used for the compression of pattern databases in combinatorial games~\cite{DBLP:conf/aaai/BreyerK10a}.

\begin{definition} 
(Perfect Hash Function)
    A \emph{perfect hash function} $h$ maps the vector $F$ of relevant features to a unique \emph{hash key} in $[0..C)$ with $C \ge |D_1| \times \ldots \times |D_k|$. 
    A minimum perfect hash function imposes $C = |D_1| \times \ldots \times |D_k|$.
\end{definition} 

Although conciseness is an issue, we do not require the hash function to be minimal and use the hash key to pack the feature vector into an (unsigned) integer datatype, which is possible if its number of bits $B$ satisfies
$\lceil \log_2|D_1| \rceil+\ldots+\lceil \log_k|D_k| \rceil \le B.$

Being an injective mapping, perfect hashing enables efficient \emph{ranking} and \emph{unranking}. 
For $k$ parameters with length
$l_i=\lceil \log_2|D_i|\rceil$ and partial sums $L_i=\sum_{j=1}^{i-1} l_i$, 
we use a bit mask $B_i = 2^{k}-1 =(1 \ldots 1)_2$ and bit operations ($\wedge$,$\gg$, $\ll$)

\paragraph{Ranking} We hash the vector or relevant features
$$F(g) = (f_1(g),\ldots,f_k(g))$$
as follows. First, we set $h(g) := f_1(g)$, and then, for each $i = 2,\ldots,k$ iteratively, we update $$h(g) := (h(g) \ll B_i) + f_i(g).$$  

\paragraph{Unranking} By virtue of the perfect hashing for the hash value
$h = h(g)$ we uniquely restore
$$f_{i}(h) := (h \wedge (B_{k-i} \ll L_{k-i})) \gg L_{k-i},  \mbox{ for } i \in \{1,\ldots,k\}.$$  
For the example of $k=7$ we have
\begin{eqnarray*}
f_1(h) &=& (h \wedge (B_7 \ll L_6)) \gg L_6 \\
f_2(h) &=& (h \wedge (B_6 \ll L_5)) \gg L_5 \\
f_3(h) &=& (h \wedge (B_5 \ll L_5)) \gg L_4 \\
f_4(h) &=& (h \wedge (B_4 \ll L_4)) \gg L_3 \\ 
f_5(h) &=& (h \wedge (B_3 \ll L_2)) \gg L_2 \\ 
f_6(h) &=& (h \wedge (B_2 \ll L_1)) \gg L_1 \\ 
f_7(h) &=& h \wedge B_1; 
\end{eqnarray*}          

As we might encounter missing values, namely feature vectors, to which we have not mapped any game in the game database(s), the hash tables might be layered e.g. into a foreground and background table, with a reduced 
set of features for the latter, so that all buckets have valid statistics. The background table is only addressed, if the lookup in the foreground table does not find any statistical value. 

As a side effect, the vectors are stored lexicographically, which means that
$h(g) \ge h(g')$ if and only if $F(g) \ge_{lex} F(g')$, where
$(x_1,\ldots,x_k) \ge_{lex} (y_1,\ldots,y_k)$ if (a) $x=y$
or (b) if $i^* = min\{i : x_i \neq y_i\}$ then $x_{i^*} > y_{i^*}$.
Therefore, we can use dictionary ordered data types to maintain
(key,value)-pairs. For every key $k$ in value $v$,
we store two numbers: the sum of all games won together with the
number of games played. When printing the ordered dictionary, we flush a table that lexicographically sorts the features.

\begin{definition} (Winning Probability)
    Assume a card game with two teams, the declarer team and the opponent team.
    The \emph{winning percentage} $p(Q)$ for question $Q$ with feature vector
    $F$ and induced hash function $h$ 
    on the game database $G$ is the number of won 
    games won in each bucket divided by the number
    of all games in the bucket, i.e., $p_{Q,G,h}(k) = win(k)/all(k)$, with
\begin{eqnarray*}
all(k) &=& |\{g \in G \mid h(g) = k\} \\
won(k) &=& |\{g \in G \mid h(g) = k\ \wedge \mbox{declarer team wins $g$} \}|    
\end{eqnarray*}
\end{definition}

In Fig.~\ref{fig:learningproc} we provide the pseudocode of the general
outer-learning procedure to update
a table based on newly played games. 
As folded games are often not recorded, we assume that the PGN information is split into input (the deals) and output (the game play information). The games in the two files are connected via the game ID. For each bucket in the hash bucket, we maintain the number of games being played and the number of games being won. The number of games played is used as confidence, so that hash buckets with a small confidence level might be neglected for the lookup, resorting to more reliable sources. Additional confidence could be added by scaling the number of games that are played and won. 

\begin{figure}[t]
\begin{tabbing}
\hspace{0.1in}     \= \hspace{0.1in}  \= \hspace{0.1in} \= \hspace{0.1in}    \kill \\
{\bf Procedure} \mbox{\em OuterLearning} \\
{\bf Input}: Files $input$, $output$ (for deals and games played) \\ \qquad \quad Functions $e(xclude)$, $f(eatures)$, $h(ash)$\\
\qquad \quad Old table $O$ (used as optional bias)  \\
{\bf Output}: New table $T$  \\
\> $T \leftarrow O$ \` \{ include old table as bias \} \\ 
\> $d \leftarrow 0$ \` \{ input-output offset \}\\
\> {\bf for} $l \leftarrow 0,\ldots,|output|-1$ \` \{ traverse file \}\\
\>\>  $o \leftarrow output_l$ \` \{ select game \} \\
\>\>   {\bf if}  $(l+d \ge |input|)$ {\bf break}  \` \{ termination \} \\
\>\>  $i \leftarrow input_{l+d}$ 
\` \{ aligned input \} \\
\>\>  {\bf if} ($o.id \neq i.id$) 
\` \{ game not played \}
\\ \>\>\>      $d\leftarrow d+1$; {\bf continue} 
\` \{ skip input game \}\\ 
\>\>  {\bf if} $(e(i,o))$ {\bf continue}; \` \{ exclusion used as filter \} \\  
\>\>    $F \leftarrow f(i,o)$ 
\` \{ get features \} \\
\>\>    $k \leftarrow h(F)$ 
\` \{ hash features to key \} \\ 
\>\> {\bf if} $(k \in T)$  \` \{ find entry \} \\
\>\>\>   $T_k.games \leftarrow T_k.games + 1$ \` \{ update \}\\
\>\>\>   {\bf if} ($o.win$)  
$T_k.won \leftarrow T_k.won+1$; \` \{ update \} \\
\>\> {\bf else} \` \{ existing entry \} \\ 
\>\>\>    $T_k.games \leftarrow 1$; \` \{ insert \}\\
\>\>\>    $T_k.won \leftarrow o.win$; \` \{ insert \}\\
\>\> $T_k.prob \leftarrow T_k.won / T_k.games$ \` \{ adapt probability \} \\
\> {\bf return} $T$ \` \{ to store new table \}
\end{tabbing}
\caption{The self-learning procedure to update statistical table on winning probabilities with optional bias.
}\label{fig:learningproc}
\end{figure}

\begin{figure}[t]
\begin{tabbing}
\hspace{0.1in}     \= \hspace{0.1in}  \= \hspace{0.1in} \= \hspace{0.1in}    \kill \\
{\bf Procedure} \mbox{\em TableReader} \\
{\bf Input}: File $input$, (for tables) \\ \qquad \quad Functions $h(ash)$ (rank)\\
\qquad \quad Old table $O$ (to be used as optional bias)  \\
{\bf Output}: Initialized table $T$  \\
\> $O \leftarrow \emptyset$ \` \{ start with empty table \} \\ 
\> {\bf for} $l \leftarrow 0,\ldots,|input|-1$ \` \{ traverse table file \}\\ 
\>\>    $F \leftarrow read(f_1,\ldots,f_l)$ 
\` \{ get features \} \\
\>\>    $k \leftarrow h(F)$ 
\` \{ hash features to key \} \\
\>\>  $O_k.won \leftarrow read()$; \` \{ insert success stats\} \\ 
\>\>    $O_k.games \leftarrow read()$; \` \{ insert game stats\}\\ 
\>\> $O_k.prob \leftarrow read()$ \` \{ insert probability \} \\
\> {\bf return} $O$ \` \{ to process new table \}
\end{tabbing}
\begin{tabbing}
\hspace{0.1in}     \= \hspace{0.1in}  \= \hspace{0.1in} \= \hspace{0.1in}    \kill \\
{\bf Procedure} \mbox{\em TablePrinter} \\
{\bf Input}: Table $T$,  
File $output$, (for tables) \\ \qquad \quad Function $h^{-1}$ (unrank)\\
{\bf Output}: File being flushed\\
\> $File.open()$ \` \{ start with empty file \} \\ 
\> {\bf for} $(k,v) \in T$ \` \{ traverse table in lex order \}\\ 
\>\> $F \leftarrow h^{-1}(k)$ \` \{ unrank key \} \\
\>\> $File.print(F = (f_1,\ldots,f_l))$ 
\` \{ print features \} \\
\>\>  $File.print(v.won)$; \` \{ insert success stats\} \\ 
\>\>  $File.print(v.games)$; \` \{ print game stats\}\\ 
\>\> $File.print(v.prob)$ \` \{ print probability \} \\
\> {\bf return} $File$ \` \{ to process new table \}
\end{tabbing}
\caption{File reader and printer routines for a chosen map}
\label{fig:readprinter}
\end{figure}

While there are different implementation options to store and retrieve the information in the tables, we preferred to employ static tables in static arrays and recompile the player each time a new table is generated. The pseudocodes for
table file reading and printing are provided in
Fig.~\ref{fig:readprinter}.

\section{Case Study}

We look at three examples applications in Skat, all of which use the above framework to answer specific questions in early stages of the game.
One reason why the winning probabilities need
to be retrieved fast is that during bidding,
skat putting, and game selection they are used
very often, e.g. in bidding for each possible game, at each bidding step for each game type and every possible Skat cards to take and discard.

\section{Winning Probabilities for Declarer}

The most important aspect for the bidding stage, discarding cards, and game declaration stage, and for the game plan in card games like Skat, is to estimate the winning probabilities for each game type, which depends on several winning features including the encountered opponent bidding strength. While being useful by the potential declarer, some of this information can
be exploited also by the opponents, reasoning about the cards of the declarer, e.g., by enumerating and evaluating the potential hands he might have.

G\"ossel~\cite{Rainer} identified nine critical winning parameters for suit games and seven winning parameters for grand games. 
If we denote $f_1,\ldots,f_l$ for the $l$ features and $h$ for the hash function, based on expert games, our objective is to address a
hash table for storing the associated winning probability. 
\[
P_w(h) = h(f_1(h),\ldots,f_l(h)).
\]
Based on more than 200 million human card games, 
base tables (one for suit and one for grand games) were built, in which the winning statistics were kept. As not each bucket was filled with data, a smaller but completely filled backup table was used for hash table misses, thereby solving the missing value problem.
The issue with the approach of is that the base tables, 
constructed in a half-automated labor-intensive process, through providing accurate winning probability estimates, are static. In this approach, it is cumbersome to include new games and weight in the play strength. In other
words, the base table does not improve with new incoming data. Especially, when
targeting AI players that exceed the human playing strength, a dynamic learning framework is needed. 

For suit games, we have the ranking function based on the winning features as proposed by \cite{Rainer}: 
\emph{numberoftrumps}, 
\emph{jacksgroup}, 
\emph{trumptenaces}, 
\emph{nontrumpaces}, 
\emph{nontrumptens}, 
\emph{lostcards}, 
\emph{biddingvalue}, 
\emph{skatvalue}, and
\emph{numberoffreesuits};
while for grand games we use the characteristics
\emph{jacksgroup}, 
\emph{numberofaces}, 
\emph{numberoftens},  
\emph{lostcards}, 
\emph{biddingvalue}, 
\emph{posplayer}, and 
\emph{skatvalue}.
As in the proposed generic approach, the
features change for different decision questions to be resolved and card games to be played, and we will not dwell on explaining them in detail.

\section{Starting Card for Opponent} 

While in Skat there are some general rules that may be refined, e.g., on playing long nontrump suits on the long way to the declarer, 
and short nontrump suits on a short way to the declarer, a more daunting question for the opponent team is whether or not a high or a low card of a chosen suit should be selected to initiate their play. 

To statistically detect the issuing most effective for the opponent, we analyze the structure
of the opponents' hands and trick-taking
information. The chosen critical parameters for suit games are: 
a) trump strength, clustered into three different values; 
b) length of the possible nontrump opening suit; 
c) is there an ace in the opening suit; 
d) is there a ten in the opening suit; 
e) table position relative to the declarer; 
f) has my partner bid for the opening suit as a trump; and 
g) {\bf in case of having an ace, was it played?}
For grand games a similar set was chosen: 
a) position of the player at the table relative to the declarer; 
b) length of the possible nontrump opening suit, 0 if jack was played; 
c) has my partner bid for the opening suit as a trump; 
d) code for constellation of Jacks; 
e) is there an ace in the opening suit; 
f) is there a ten in the opening suit; 
and g) {\bf in case of having an ace, was it played?}  

Unlike the previous case used for bidding skat putting and game selection, here we look also at
trick-taking information in the corpus of
suit and grand games.
 
Although different questions can be resolved in
this matter, the \emph{Under-Ace} Question $U$, i.e., whether
or not an ace in a nontrump color
should be issued or not. As \emph{Under-Ace} is a winning feature and both cases (playing under ace or not) are contained in two different hash buckets, $U$ can be answered by comparing the statistics of winning.

\section{Null Game Winning Probabilities}

The winning probabilities for null games
follow the suggestion of~\cite{Lasker,Lasker2}, namely
that the winning probability for the
declarer's hand can be approximated by
the product of the winning probabilities in
each suit, leading to tables with 256 entries. 

We combine statistics on the winning probabilities of a hand $h$ in each suit as follows.
\[ P_w(h) = \prod_{suit \in 
\{\clubsuit,\spadesuit,\heartsuit,\diamondsuit\}}
%\{club,spade,heart,diamond\}} 
P_w(h,suit), \]
with suit probabilities
$P_w(h,suit)$, one for each variant of the Nullspiel. We implemented refinements that take
into account a) the type of Null game 
(standard, hand, ouvert, ouvert hand), b)
if it is the declarer to move (in first trick), 
and c) the number of discarded Skat cards 0,1, or 2 discarded in the
chosen suit. With the above framework,
refined winning probabilities can be deduced for a
large set of games, and winning parameters have
been added. 

To avoid interaction, we restricted the data mining analysis of games to ones where only one
weakness appears, and deduce if the declarer has been
hit in this color. We also record the card that was chosen to beat the declarer (the issuing card in the last trick).

\section{Experiments}

We ran our experiments on an
AMD EPYC 9634 with 16 cores, evaluating
the player compiled with gcc-version 12.2.0 (Debian 12.2.0-14) and optimized via
\verb/-O3/. Although each batch of games runs on 1 core, we ran 20  independent batches in parallel. 
For this, we generated and stored 100 million random deals with the Mersenne Twister, of which the first 30 million were used. The table position for issuing the first card was changed clockwise.

Evaluating the strength of an AI player for card games is tricky, as the declarer strength might reflect the opponent
weakness and vise versa, so that scoring one or the other can be misleading. Even optimizing the code for maximum (Seeger-Fabian) score values that include benefits for games won by the opponents can be misleading.
Therefore, the evaluation
criterion we are mainly interested in is the accuracy, namely, how well the player matches the result predicted by the open-card solver. We reuse a set of 83,844 recorded human games whose deals were not completely random, as folded games, i.e., games without bids, were no longer present. When splitting the criterion into two, we have the true positive rate and the false negative rate, meaning that the open-card solver predicts a win (loss) and the outcome was a win (loss). 

Each evaluation of 88,844 games took about 2.5h on our machine. The automated generation of all tables, even with 30 millions of games in the database, was available in less than four minutes, including about a minute to recompile the player. Hence, for 100 million games, we would need less than 15 minutes to construct and integrate all cumulative tables. The hyper-parameters for the search functions (paranoia search and end game analyzes) were set, so that on average 1-3 seconds per computer game was needed, as required for swift human-computer interactive play. Self-playing one million games on 16 cores, thus, equates to about 1.5 days, so that generating 30 million AI games took more than one month. 

\setlength{\tabcolsep}{2pt}

\begin{table}[t]
%\begin{small}
\begin{tabular}{ccccccc}
Game&OCS&AI&Count&Total&\%Games& Acc \\ \hline
Suit & 0&0&1094011& \\
Suit & 0&1&980015& \\
Suit & 1&0&157344& \\
Suit & 1&1&4011994&6243364&64.4\%&81.8\% \\ \hline
Null23/35 & 0&0&156112& \\
Null23/35 & 0&1&165322& \\
Null23/35 & 1&0&2463& \\
Null23/35 & 1&1&142154&466051&4.8\%&64.0\% \\ \hline
Grand & 0&0&86435& \\
Grand & 0&1&181883& \\
Grand & 1&0&34793& \\
Grand & 1&1&2283658&2586769&26.7\%&91.6\% \\ \hline
Null46/59 & 0&0&39001& \\
Null46/59 & 0&1&729& \\
Null46/59 & 1&0&2486& \\
Null46/59 & 1&1&362465&404681&4.2\% & 99.2\% \\ \hline
Total & & & & 9700865&9700865&100.0\% \\ \hline
Folded & & & & & 299135 & 2.99\% \\
\end{tabular}
%\end{small}
\caption{Distribution of 10 million games in 
self-play of three identical Skat engines w.r.t. game type and predicted outcome of open-card solver. }\label{fig:distribution}
\end{table}

\begin{figure}
    \centering
\includegraphics[width=1\linewidth]{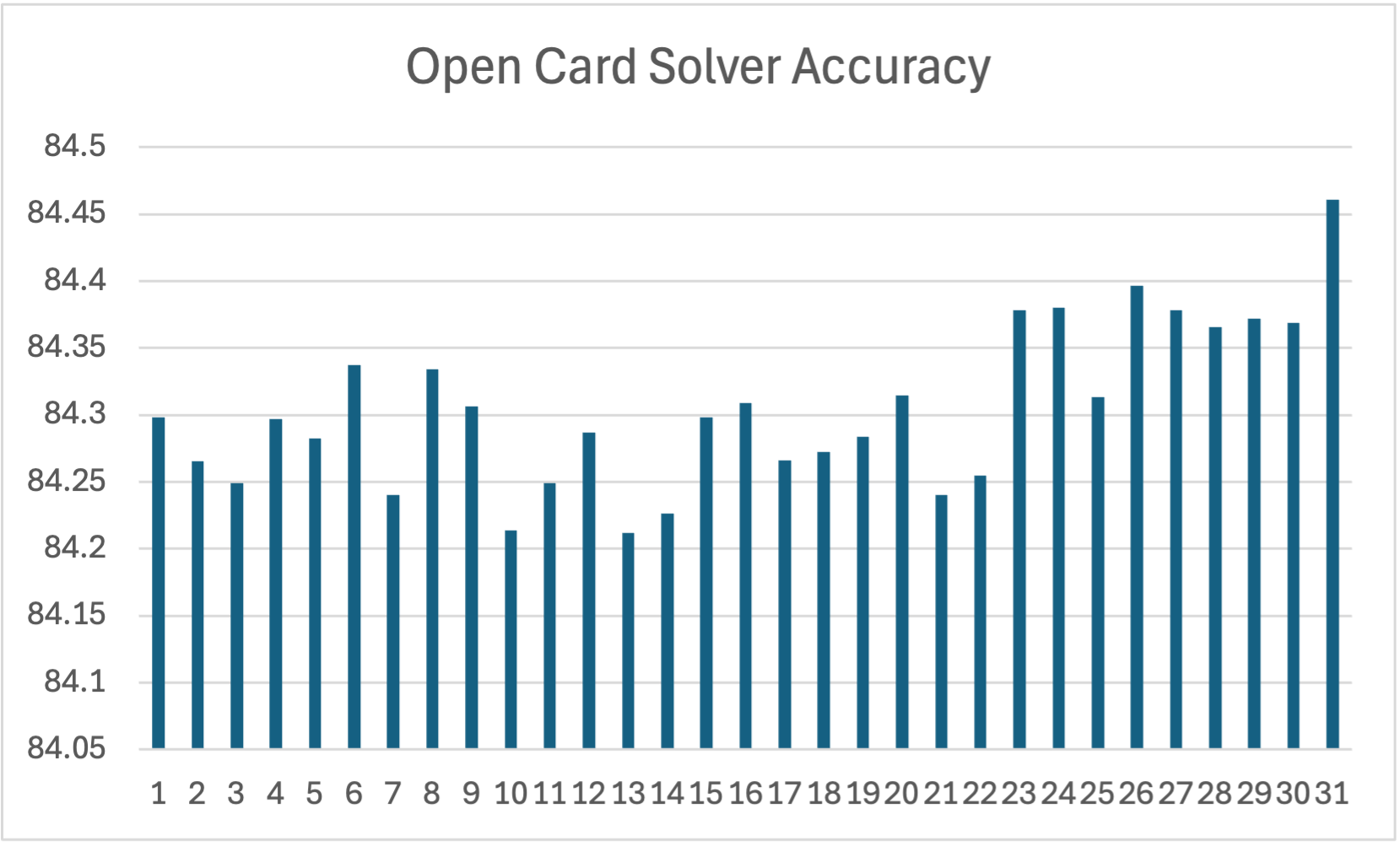}
    \caption{Change in accuracy values during the outer-learning framework along 
 playing 31 million Skat games of three AIs, with player updates after 1-5, 10, 20, 22, and 30 million games.}   \label{fig:accuracy}
\end{figure}

Figure~\ref{fig:accuracy} show the
change in the accuracy and winning percentage values during the outer-learning process. Tables have been recomputed after 5, 22 and 30 
million games. While for 5 million games the picture the improvement is not as clear, we definitely see an increase in these values after 22 million and 30 million games, respectively. 

\begin{figure}
    \centering
\includegraphics[width=1\linewidth]{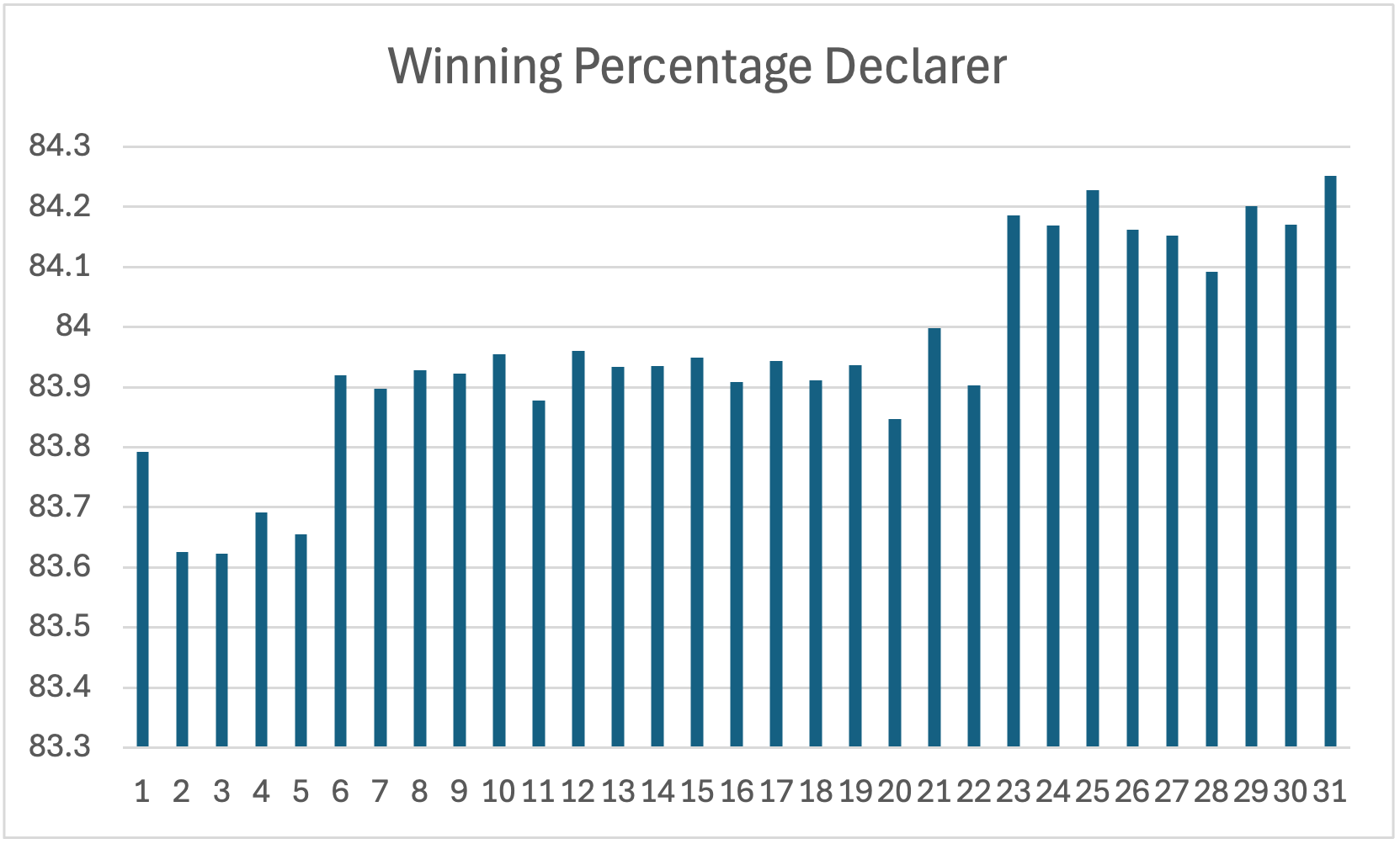}
    \caption{Change in winning percentage values during the outer-learning framework along playing 31 million Skat games of three AIs, with player updates after 1-5, 10, 20, 22, and 30 million games.}
    \label{fig:winning}
\end{figure}

It has been shown \cite{DBLP:conf/cg/Edelkamp22} that a Skat AI can beat a top human player in a regular match of 10 series of 36 games, thus with a statistically relevant number of around 360 games, but admitted that top humans may still have an edge as the volatility in games is still large. Closing the remaining gap remains a challenge. Edelkamp  reported an accuracy of $84.5\%$ replaying the chosen set of $83,844$ human games (with AI bidding and game selection), where the human players only achieved an accuracy of $76.7\%$.
%For constructing the base table, manual human invention and half-manual interpolation were used.

\begin{table}[t]
\begin{center} 
\begin{tabular}{ccccc}
Human&OCS&AI&\#Games($20\cdot{10^6}$) & \#Games($30\cdot{10^6}$) \\ \hline
 0&0&0&  5113 & 5008 \\    
0&0&1&  3270 & 3294\\    
0&1&0&   361 & 368 \\    
0&1&1&  4673 & 4722\\    
1&0&0&  6337 & 6084 \\    
1&0&1&  7699 & 7690\\    
1&1&0&  1269 & 1195\\   
1&1&1& 53812 & 54081 \\ \hline
Total &&& {\bf 82534} & 82442 \\
Accuracy &&& 84.73 & {\bf 84.78} \\
\end{tabular} 
\end{center}
\caption{Replaying 83,844 Human games with 
learning tables on 22 and 30 million games. The values in the first three columns denote a win (1) or a lost (0) for the human in the original game, the open-card solver, and the AI in the replay.} \label{fig:replay5}
\end{table}

\begin{figure}[t!]
    \centering \includegraphics[width=1\linewidth]{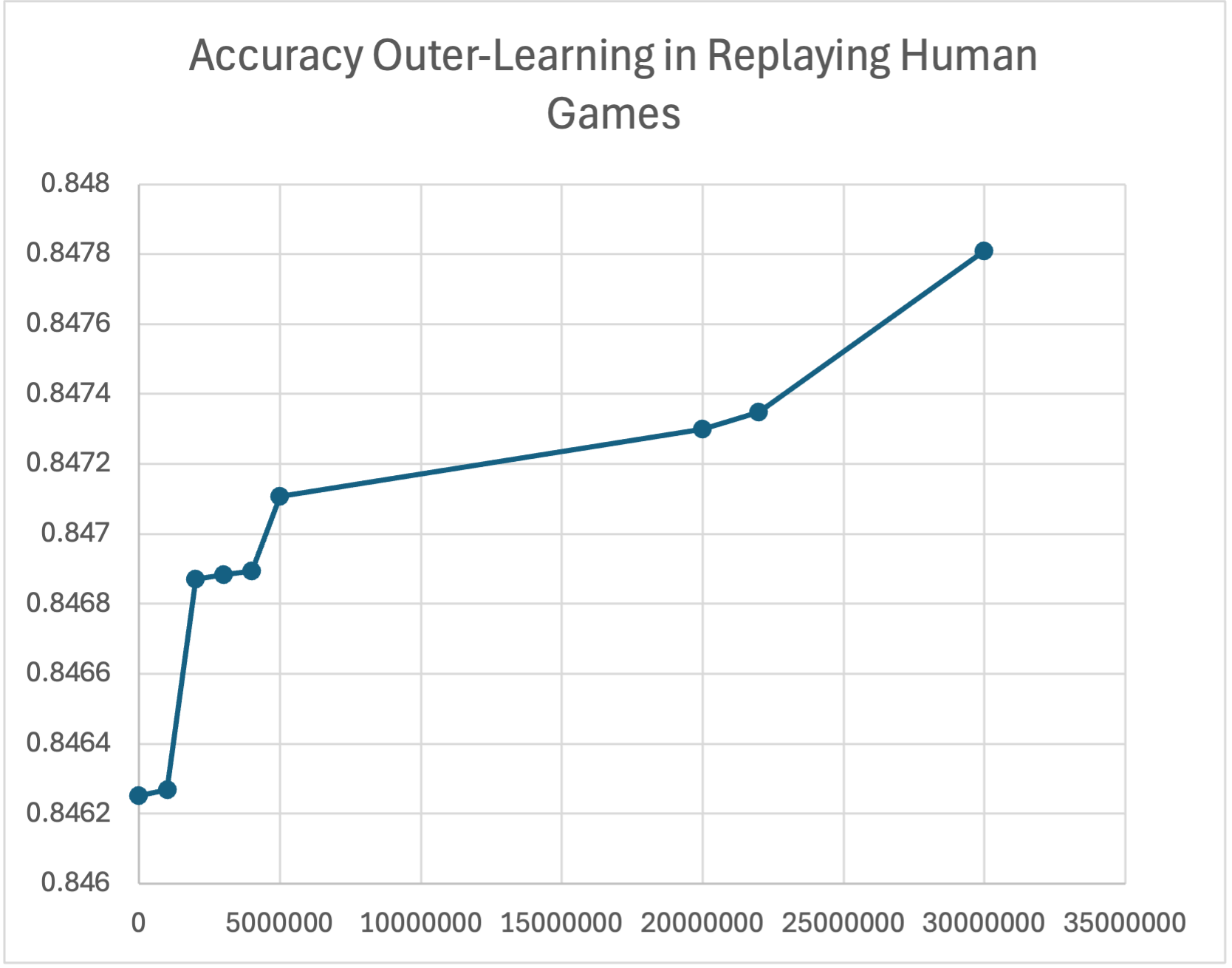}
    \caption{Accuracy for replaying 83,844 Human games plotted over the outer-learning process on the 30 million games (at each data point in the plot, new tables were generated).}
    \label{fig:evolution}
\end{figure}

%\begin{figure}[t]
%    \centering
%    \includegraphics[width=0.991\linewidth]{seeger.png}
%    \includegraphics[width=0.991\linewidth]{win.png}
%    \includegraphics[width=0.991\linewidth]{games.png}
%    \caption{Other performance indicators.}
%    \label{fig:performance}
%\end{figure}

Table~\ref{fig:replay5} shows that for
the learning framework
the accuracy increased to $84.73\%$ at around 20 million 
and $84.78$ at 30 million games.
Although the improvement in accuracy of $0.23\%-0.27\%$ compared to the value given in \cite{DBLP:conf/cg/Edelkamp22} appears small at first glance, the increase in playing strength will turn out to be significant. 
%For the case of learning with about %30 million games, this can be argued by considering the reduction $2\%$ of the remaining games that do not match the open card solver recommendation from $12.5\%$ to $12.25\%$. 

Fig~\ref{fig:evolution} shows the evolution of the accuracy value in the test benchmark during the learning process over the first 30 million games. We highlight the changes when newly inferred tables were used for the evaluation. The improvement in accuracy is larger in the beginning but still noticable later on. %Fig.~\ref{fig:performance} shows the values and other performance indicators during learning.

We duplicated the learning by reinserting the
result of the training. This led to a decrease in the accuracy to $84.70\%$, so it was not paying off. 
We also looked at zero learning, i.e., using
no base table at all. At 22 million games, the
grand table shrunk from 113066 to 18789 entries, and
the suit table from 261389 to 62404. However, the precision fell to 83. 76\%, showing
that a base tables (inferred in more than 200 human games) is still needed for the top play.

While the improvement in the accuracy value look small on first sight, we evaluated the learning effect of outer learning in a direct competition. 
Table~\ref{tab:thousand} provides
the results of the play on 1000 random deals played with differently assigned seats on our Skat server. The depicted score values are the widely accepted Seeger-Fabian score\footnote{see https://en.wikipedia.org/wiki/Skat\_scoring}. 
As the results in the smaller experiment may
not be conclusive, Table~\ref{tab:severalthousand} extended the first experiment to several thousand games, and Table~\ref{tab:sum} displays the accumulated total of the two experiments and illustrates clear dominance of the competitive fully fledged play on random deals with outer learning over the one without. We validated this finding in a number of additional experiments.

\begin{table}[]
    \centering \small
    \begin{tabular}{c||c|c|c}
    \multicolumn{4}{c}{two bots with OL, one without} \\ \hline
     Exp.~1  & +OL
       & +OL
       & -OL \\ \hline\hline
    won & {\bf 292} & 275 &  256 \\
    lost & 56 & {\bf 42} & 48 \\
    score & 25788 & {\bf 26580} & 23947 \\    
    \end{tabular}
 \qquad   
    \begin{tabular}{c||c|c|c}
    \multicolumn{4}{c}{one bot with OL, two without} \\ \hline
      Exp.~1 & +OL
       & -OL
       & -OL \\ \hline\hline
    won & {\bf 269} & 265 & {\bf 269} \\
    lost & {\bf 38} & 49 & 49 \\
    score & 26737 & 23895 & {\bf 27268} \\    
    \end{tabular}
    \caption{Experiment 1 of server play competition (2:1) and (1:2) between bots with (+) and without (-) outer learning (OL) on 1000 AI games.}
    \label{tab:thousand}
\end{table}

\begin{table}[]
    \centering \small
    \begin{tabular}{c||c|c|c}
    \multicolumn{4}{c}{two bots with OL, one without} \\ \hline
      Exp.~2 & +OL
       & +OL
       & -OL \\ \hline\hline
    won & 2208 & {\bf 2256} &  2158 \\
    lost & 428 & {\bf 394} & 400 \\
    score & 201133 & {\bf 211381} & 201681 \\    
    \end{tabular}
   \qquad   
    \begin{tabular}{c||c|c|c}
    \multicolumn{4}{c}{one bot with OL, two without} \\ \hline
       Exp.~2& +OL
       & -OL
       & -OL \\ \hline\hline
    won & {\bf 1999} & 1942 & 1944 \\
    lost & {\bf 345} & 357 & 349 \\
    score & {\bf 189211} & 176572 &  180795 \\    
    \end{tabular}
    \caption{Experiment 2 of server play comparison between bots with (+) and without (-) outer learning (OL) on 7127 (1:2) and 8080 (2:1) AI games.
    %(the experiment on 10000 games was terminated externally.) 
    }
    \label{tab:severalthousand}
\end{table}

\begin{table}[]
    \centering \small
    \begin{tabular}{c||c|c|c}
    \multicolumn{4}{c}{two bots with OL, one without} \\ \hline
      Total & +OL
       & +OL
       & -OL \\ \hline\hline
    won & 2500 & {\bf 2531} &  2414 \\
    lost & 484 & {\bf 436} & 448 \\
    score & 226921 & {\bf 237961} & 225628 \\    
    \end{tabular}
 \qquad   
    \begin{tabular}{c||c|c|c}
    \multicolumn{4}{c}{one bot with OL, two without} \\ \hline
       Total& +OL
       & -OL
       & -OL \\ \hline\hline
    won & {\bf 2268} & 2207 & 2213 \\
    lost & {\bf 383} & 406 & 398 \\
    score & {\bf 215948} & 200467 &  208063 \\    
    \end{tabular}
    \caption{Sum of Experiments 1 \& 2 of server play comparison between bots with (+) and without (-) outer learning (OL) on 8127 (1:2) and 9080 (2:1) AI games. }
    \label{tab:sum}
\end{table}

\section{Conclusion}

In fully observable board games, there has been considerable success with deep learning based on bootstrapping game databases using neural networks for move prediction. These approaches applied to Chess, Shogi or Go, start from a series of expert games, or in learning from mere self-play. However, so far, deep neural networks have shown only marginal impact in complex card games such as Bridge or Skat~\cite{sychrovský2025approximatingnashequilibriageneralsum}. 
Although there are machine learning procedures that have been applied to the bidding stages of card games, e.g. applying game theory~\cite{DBLP:conf/ijcai/BuroLFS09},and other algorithms that were effective in trick taking~\cite{DBLP:journals/corr/abs-1911-07960}, a general learning framework that was effective for the initial stages of complex card games was lacking.

The proposed outer-learning approach is the first to fully automate bootstrapping for continuous learning in trick-taking card games, enabling recursive improvement via AI self-play and data merging. It successfully includes bootstrapping to derive statistics and improve the play of trick-taking card games completely automatically. The approach relies on a statistical selection of winning features and that, once established, can be applied to many early decision choice points of a card game. It may even apply to other turn-based incomplete information games that provide the outcome of two competing and otherwise cooperating teams.

With this paper, we provided a general framework showing that it is possible to apply bootstrapping to improve the AI card players' performance over time, during play. It applies to the determination of winning probabilities and asking particular questions on the statistics of certain decisions early in the game. The programmer has to provide features that are relevant for the answer, and puts games into buckets of comparable strength. A compromise between the
number of games and the number of buckets must be found to provide sufficient data for the statistics.

\begin{table}[t!] 
\centering
\begin{tabular}{cc}
pos & content \\ \hline
1  &         ID \\
2  &         game\\
3  &         position declarer in {0..2} \\
4..6 &       bidding value for players 0..2 \\
7  &         hand 0/1 \\
8  &         Schneider 0/1 \\
9  &         Schneider announced 0/1 \\
10 &         Schwarz 0/1 \\
11 &         Schwarz announced 0/1 \\
12 &         Ouvert 0/1 \\
13 &         declarer won 0/1 \\
14 &         folded 0/1 \\ 
15 &         contract level 1..11,-1..-11 \\
16 &         eyes declarer \\
17..19 &     hand strength value 0..10 for players 0..2 \\
20..29 &    hand player 0 \\
30..39 &    hand player 1 \\
40..49 &    hand player 2 \\
50..51 &    Skat cards taken \\
52..53 &    Skat cards put \\
54..56 &    cards 1st trick \\
57      &    position 1st trick winner \\  
58      &   1st trick eyes ($>$0 for declarer, $<$0, else) \\
59..61 &    cards 2nd trick \\ 
62      &    position 2nd trick winner \\
63      &    2nd trick eyes ($>$0 for declarer, $<$0, else) \\
\vdots  &    \vdots
\end{tabular}
\caption{PGN for Skat; -1 for a folded game, 0..3 for a Suit  $(\clubsuit,\spadesuit,\heartsuit,\diamondsuit)$, 4 for a Grand, 5 for a Null, 6 for a Null Ouvert, 7 for a Null Hand, 8 for a Null Ouvert Hand.}
\label{tab:pgn}
\end{table}

%An automated extraction of features has not yet been determined, but we follow the guidelines of G\"o{\ss}l to validate them by analyzing game data. 
%We also aim to include the impact of a game with player strength and to develop a strategy for discarding older games.

\paragraph{Acknowledgements}

We thank Rainer Gößl for his collaboration and the idea of the implementation, and Daniel Schäfer for the Skat app screenshot. The contribution was supported by GA\v{C}R project
24-12046S.

\clearpage
\pagebreak
\vfill
\pagebreak

\bibliography{bibliography}
\end{document}